# Creating Pro-Level AI for a Real-Time Fighting Game Using Deep Reinforcement Learning

**Inseok Oh\*, Seungeun Rho\*, Sangbin Moon, Seongho Son, Hyoil Lee and Jinyun Chung†**

NCSOFT, South Korea

{ohinsuk, gloomymonday, sangbin, hingdoong, onetop21, jchung2050}@ncsoft.com

## Abstract

Reinforcement learning combined with deep neural networks has performed remarkably well in many genres of games recently. It has surpassed human-level performance in fixed game environments and turn-based two player board games. However, to the best of our knowledge, current research has yet to produce a result that has surpassed human-level performance in modern complex fighting games. This is due to the inherent difficulties with real-time fighting games, including: vast action spaces, action dependencies, and imperfect information. We overcame these challenges and made 1v1 battle AI agents for the commercial game "Blade & Soul". The trained agents competed against five professional gamers and achieved a win rate of 62%.

This paper presents a practical reinforcement learning method that includes a novel self-play curriculum and data skipping techniques. Through the curriculum, three different styles of agents were created by reward shaping and were trained against each other. Additionally, this paper suggests data skipping techniques that could increase data efficiency and facilitate explorations in vast spaces.

Since our method can be generally applied to all two-player competitive games with vast action spaces, we anticipate its application to game development including level design and automated balancing.

## Introduction

Reinforcement learning (RL) is extending its boundaries to a variety of game genres. In PVE (player versus environment) settings, such as those found in Atari 2600 games, RL agents have exceeded human level performance using various methods (Mnih *et al.* 2015; Mnih *et al.* 2016; Schulman *et al.* 2017; Hessel *et al.* 2018). Likewise, in PVP (player versus player) settings, neural networks combined with search-based methods beat the best human players in turn-based two player games—such as Go and Chess (Silver *et al.* 2018b). Recently, RL research in games has shifted focus to the PVP settings found in more complex video games such as StarCraft2 (Vinyals *et al.* 2017), Quake3 (Jaderberg *et al.* 2018), and Dota2 (OpenAI 2018).

Fighting games—as one of the most representative types of complex PVP games—have been the focus of multiple studies that have made progress in this area. For instance, MCTS based methods (Yoshida *et al.* 2016; Kim *et al.* 2017; Ishihara *et al.* 2019) have been applied to "FightingICE(FICE)", a game platform made for the Fighting Game AI Competition (Lu *et al.* 2013). However, it is hard to fulfill real-time conditions when applied to heavier modern game engines with longer query times. Additionally, a deep RL based agent (Li *et al.* 2018) was trained against a rule-based fixed opponent in "Little Fighter 2(LF2)". However, since the opponent's decision is unknown at a player's decision time, agents trained against rule-based AIs cannot be generalized for unseen opponents. Our approach is largely similar to that of Firoiu *et al.* (2017) in which a self-play deep RL method was applied to "Super Smash Bros. Melee (SSBM)". However, the complexity of state and action space is significantly limited compared to our 3D environment with complex game rules. We created pro-level AI agents for the real-time fighting game "Blade & Soul (B&S) Arena Battle" via novel self-play based reinforcement learning.

|      | Year | Commercial | Dimension | Pro-scene |
|------|------|------------|-----------|-----------|
| FICE | 2013 | X          | 2D        | X         |
| LF2  | 1999 | O          | 2.5D      | X         |
| SSBM | 2001 | O          | 2D        | O         |
| BAB  | 2013 | O          | 3D        | O         |

Table 1: Fighting games from other works

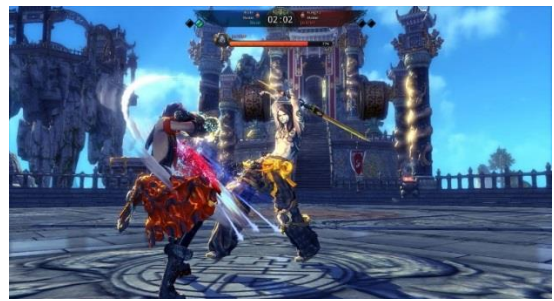

Figure 1. A scene from the B&S Arena Battle

B&S is a commercial massively multiplayer online role-playing game. It supports duels between two players called "B&S Arena Battles (BABs)". As presented in Table 1, BAB is a more modern fighting game compared to the games considered in other works; hence, it has much more complex game dynamics and heavier game engines. Additionally, a large number of people play BAB and it has more active professional scenes [1] than other fighting games. BAB's larger number of active professional scenes stands out more significantly when compared to FightingICE, which was designed solely for research purposes.

Figure 1 displays a scene from BAB. BAB is a two-player zero sum game. In BAB, two players fight against each other to reduce their opponent's HP (health point) to zero within three minutes. To master BAB, an agent must be able to deal with multiple challenges.

First, an agent must manage vast action and state spaces. An agent must make skill, move, and targeting decisions simultaneously, which yields many possible combinations. As a rough estimate, there are 144 potential actions for each time step: 8 (avg. # of avail. skills) * 9 (8 directional + no move) * 2 (facing opponent or moving direction). Since the average game length is 1200 time steps (120 s), numerous scenarios are possible—not considering the opponent's actions.

Moreover, an agent must consider the dependencies between skills: e.g., a skill may become available only for a short period of time following the use of another skill. As a result, out of the 45 skills in total (including "no-op"), the set of skills available at a given time constantly changes. The agent must also consider the properties of each skill because they have different cooldown times (required interval for re-using a skill) and SP (skill point) consumptions, and serve one or more of five different functions: damage dealing, crowd control (which functions to make the opponent incompetent; abbreviated CC), resistance (which functions to make the player immune or resistant to CC skills), escape, and dash.

Lastly, an agent must deal with imperfect information settings. Because BAB is a real-time game, two players make their decisions simultaneously. This indicates that an agent is required to make decisions without knowing the opponent's decision or strategy. Hence, BAB can be considered to be a series of rock-paper-scissors games. For example, when a player uses a resistance skill and the opponent uses a crowd control skill at the same time, the player gains advantage over the opponent. As a result, the essence of the problem is to approximate a Nash equilibrium strategy so that the agent can respond appropriately to any opposing strategy.

To tackle these challenges, we have made improvements to vanilla self-play algorithm by diversifying opponent pools and skipping data to facilitate exploration. The main contributions of this work are as follows:

- We devised a novel self-play curriculum with agents of different styles. The curriculum made these agents compete against each other and reinforced the agents simultaneously, rendering the agents capable of handling a variety of opponents. We empirically demonstrate that our curriculum outperforms vanilla self-play method.
- We diversified the fighting style of the game-playing AIs by reward shaping (Ng *et al.* 1999). We created three types of agents with different fighting styles: aggressive, defensive, and balanced. We anticipate its application to game development including level design and automated balancing.
- We introduced data skipping techniques to enhance exploration in vast space. These can be generally applied to any two-player real-time fighting games.
- We evaluate our agents by pitting them against professional players in the 2018 B&S World Championship Blind Match. Our AI agents won three out seven matches, while the aggressive one beating all professional players both in the live event and pre-test.

## Background

### Reinforcement Learning

In reinforcement learning (Sutton and Barto 1998), agent and environment can be formalized as a Markov decision process (MDP) (Howard 1960). For every discrete time step $t$, an agent receives a state $s_t \in S$ and sends an action $a_t \in A$ to the environment. Then, the environment makes a state transition from $s_t$ to $s_{t+1}$ with the state transition probability $P^a_{ss'} = P[s'|s,a]$ and gives a reward signal $r_t \in \mathbb{R}$ to the agent. Therefore, this process can be expressed with $\{S, A, P, R, \gamma\}$, where $\gamma \in [0,1]$ is a discount factor, which represents the preference for immediate reward over long-term reward. Here, the agent samples an action from a policy $\pi(a_t|s_t)$, and the learning process modifies the policy to encourage good actions and suppress bad actions. The objective of the learning is to find the optimal policy $\pi^*$ that maximizes the expected discounted cumulative reward.

$$\pi^* = argmax_\pi E_\pi[\Sigma \, r_t * \gamma^t]$$

---

[1] 9 regional league winners from all over the world (including KOR, NA, EU, RUS, and CHN) participated in the 2018 B&S world championship (fourth annual event). The winning prize was approx. $50k (compared to Tekken7: $30k)

## Real-Time Two Player Game

In a real-time two player game, there are two players, namely, the agent and the opponent. Both of them send an action to the environment at the same time. Let us denote the policy of the agent as $\pi^{ag}$, and the policy of the opponent as $\pi^{op}$. Each samples an action from its own policy for every time step.

$$a_t^{ag} \sim \pi^{ag}(a_t^{ag}|s_t), \; a_t^{op} \sim \pi^{op}(a_t^{op}|s_t)$$

Then, the environment makes a state transition by considering those two actions jointly.

$$s_{t+1} \sim P(s_{t+1}|s_t, a_t^{ag}, a_t^{op}), \; r_{t+1} = R(s_t, a_t^{ag}, a_t^{op})$$

Here, the MDP can be expressed as $\{S, A^{ag}, A^{op}, P, R, \gamma\}$. If $\pi^{op}$ is fixed, then we can regard the opponent as a part of the environment by marginalizing the policy of the opponent.

$$P'(s_{t+1}|s_t, a_t^{ag})$$
$$= \sum_{a_t^{op}} \pi^{op}(a_t^{op}|s_t) * P(s_{t+1}|s_t, a_t^{ag}, a_t^{op})$$

Then, the MDP expression turns into a simpler form with $P'$: $\{S, A^{ag}, P', R', \gamma\}$. This expression is coherent with the one player MDP. Therefore, any methods for the original MDP work in this form as well. However, $\pi^{op}$ is not fixed in general, and our agent does not know which $\pi^{op}$ it is going to face. We propose a self-play curriculum with diversified pool of $\pi^{op}$ in the following section.

## BAB as MDP

If we assume $\pi^{op}$ or the pool of $\pi^{op}$ is fixed, BAB can be expressed as an MDP. Figure 2 illustrates the agent-environment framework in BAB. LSTM (Hochreiter and Schmidhuber 1997) based agents interact with the BAB simulator, which acts as the environment. For every time step with 0.1 sec intervals, state $s_t$ is constructed from the history of observations $H_t = \{o_1, o_2, \ldots, o_t\}$. To be specific, $s_t$ is composed of any information that a human can access during a game, such as HP, SP, distance from opponent, distance from the arena wall, current position, remaining game time, remaining cooldown times for all 44 skills, an agent's status info (midair, stun, down, kneel, etc.), and so on. Then, the agent decides on an action $a_t = (a_t^{skill}, a_t^{move,target})$ for every time step. Note that the targeting action space was originally continuous. We discretized it into two (facing opponent or current moving direction) and jointly considered it along with the decision to move.

Following this, the action is then sent to the environment and a state transition occurs accordingly. Here, exact rewards should also be determined. Rewards are closely related to high performance in BAB. We provided $r_t^{WIN}$, which is the reward for winning a game, and $r_t^{HP}$, the reward for the changes in HP margin. These rewards are designed based on the assumption that the more a player wins, and with more remaining HP, the better that player's performance is. $r_t^{WIN}$ is given at the terminal step of each episode with +10 for a win and -10 for a loss. $r_t^{HP}$ may occur at every time step when the agent deals damage to the opponent and vice versa. Since HP is normalized to [0, 10], $r_t^{WIN}$ and $r_t^{HP}$ have the same scale.

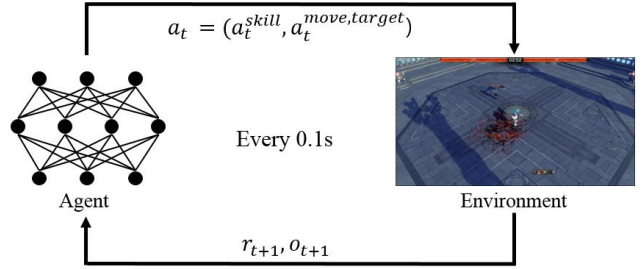

Figure 2. Agent-environment plot in BAB

$$r_t = r_t^{WIN} + r_t^{HP}$$
$$r_t^{HP} = (HP_t^{ag} - HP_{t-1}^{ag}) - (HP_t^{op} - HP_{t-1}^{op})$$

These are fundamental rewards, and additional rewards for guiding battle styles are described in the next section. The value of $\gamma$ is set to 0.995, which is close to 1.0, since all episodes in BAB are forced to terminate after 1,800 time steps (= 3 min).

## Self-Play Curriculum with Diverse Styles

Existing self-play methods (Silver *et al*. 2017; Silver *et al*. 2018a) generally use opponent pools for training. Parameters of a network are stored at regular intervals during training to create a pool of past selves. Opponents are then sampled from this pool.

Although the self-play method of RL offers a way to learn the Nash equilibrium strategy (Heinrich and Silver 2016), high coverage of strategy space is essential to efficiently find one. Vanilla self-play alone does not guarantee enough coverage for games with large problem spaces. To tackle this problem, AlphaStar (Vinyals *et al*. 2019) diversified the opponent pool by imitating different human strategy and introducing three types of agents with different match making scheme. The Poker AI, Pluribus (Brown and Sandholm 2019), hand-tuned three different strategies on top of basic blueprint strategy. The three strategies are biased toward raising, calling, and folding respectively.

Concurrently, we devised a novel self-play curriculum. We enforced diversity of agents' strategies by introducing a range of different battle styles, and agents of different styles were made to compete against each other.

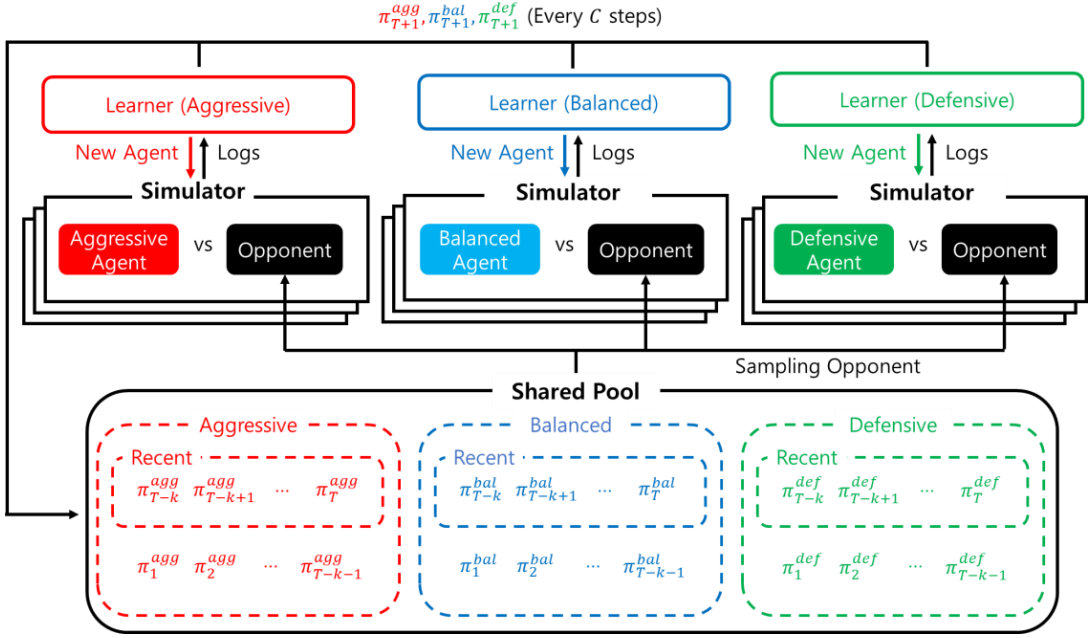

Figure 3. Overview of self-play curriculum with three different styles

## Guiding Battle Styles through Reward Shaping

One of the most noticeable fighting styles to invest with is the degree of aggressiveness. We used three dimensions of rewards to control the degree of aggressiveness. The first dimension is the "time penalty". The aggressive agent receives larger penalties per time step, and this motivates it to finish the match in a shorter period of time. The second dimension is the relative importance of the agent's HP to the opponent's HP. Aggressive players will try to reduce the opponent's HP rather than preserving their own HP, while defensive players tend to act the opposite way. The final dimension is the "distance penalty". Defensive players tend to ensure a certain distance from their opponents to respond appropriately against attacks, while aggressive players tend to approach their opponents and attack relentlessly. To realize these properties, the aggressive agent received larger penalties in proportion compared to the distance between itself and its opponent. The specific reward weights used for each style are shown in Table 2. Note that each of these three dimensions can take continuous values. This means that it is possible to create a spectrum of different fighting styles with varying degrees of aggressiveness. However, to effectively demonstrate the viability of this method, we limited the number of fighting styles to three. By using any type of additional reward signals along with $r_t^{WIN}$ and $r_t^{HP}$, this method could be applied to other fighting games in general to create agents with various fighting styles.

## Our Self-Play Curriculum

Figure 3 shows an overview of the proposed self-play curriculum with three different types of agents. Agents of each style have their own learning process, and all three agent types were trained in a concurrent manner.

Each learning process consisted of a learner and multiple simulators. The learner and the simulators work asynchronously. In the simulators, an agent constantly plays matches against randomly sampled opponents from the shared pool. The most recent $k$ models of each style are uniformly selected with total probability mass of $p$, while other models are chosen uniformly with probability $1-p$. As training goes on, $p$ is linearly annealed from 0.8 to 0.1. A higher $p$ assists in swift adaptation to the latest opponents, while a lower $p$ stabilizes the learning process by alleviating catastrophic forgetting. Each simulator sends a match log to the learner at the end of every match and updates its agent with the latest parameters received from the learner. The same procedure continues to be used through subsequent games.

The learner trains its agents in an off-policy manner using logs gathered from multiple simulators and sends the latest network parameters to the simulators on request. In addition, the learner sends its network parameters to the shared pool every C steps (e.g. C=10,000) of update. Thus, the pool has varying policies that come from the different learning processes of the different styles. These sets of model parameters are provided as opponents to each learning process. By sharing a pool, every learning agent encounters opponents of every style during training and learns how to deal with them. Therefore, agents trained via our self-play curriculum can ultimately learn how to face opponents with varying fighting styles while maintaining their own battle styles.

|  | Aggressive | Balanced | Defensive |
|---|---|---|---|
| Time penalty | 0.008 | 0.004 | 0.0 |
| HP ratio | 5:5 | 5:5 | 6:4 |
| Distance penalty | 0.002 | 0.0002 | 0.0 |

Table 2: Reward weights of each style

## Data Skipping Techniques

In this section, we detail data skipping technique, which refers to the process of dropping certain data during training and evaluation procedures.

### Discarding Passive "No-op"

In fighting games, using skills generally consumes resources, such as SP and cooldown time. Therefore, if a player overuses a certain skill, it will not be available for use during actual times of need. Thus, players should strategically use and retain their skills to ensure their availability when needed. To take this aspect into account, we concatenated a "no-op" action to the output of the policy network, allowing the agent to choose "no-op" and do nothing for a certain period if necessary. This means that our action space has 44 skills, plus an additional "no-op" action. This is significant because human play logs of BAB show that "no-op" actions take up the largest portion of skill usage among human players.

"No-op" decisions can be categorized as passive and active use cases. The passive use of "no-op" implies that an agent chooses "no-op" because there is no skill available for use. For example, when an agent is out of resources or is hit by an opponent's CC skill, an agent has no option but to choose "no-op". The active use of "no-op" means that an agent selects "no-op" strategically, even though other skills are available for use.

We discarded passive "no-op" data from both the training and evaluation phases because passive "no-ops" are not used deliberately by an agent. In addition, the method enables LSTM to reflect representations of longer time horizons because the data is not provided to the network. We show in the experiment section that skipping passive "no-ops" greatly improves learning efficiency. Note that this methodology is generally applicable to other domains where a set of available skills changes constantly and the "no-op" action is a valid option to choose.

### Maintaining Move Action

Although a single skill decision can have a substantial influence on the subsequent states, the effect of a single move decision is relatively limited. The reason is that the distance a character moves in a single time step (0.1 s) is very short considering its speed. In order for any moving decision to have a meaningful effect, the agent should make the same moving decision consecutively for several ticks in a row. This allows the agent to literally "move" and leads to changes in subsequent states and rewards. Therefore, it is difficult to train a move policy from the initial policy with random move decisions. Since the chance of a random policy making the same decision consecutively is very low, exploration is extremely limited. We therefore propose maintaining the move decision for a fixed number of time steps.

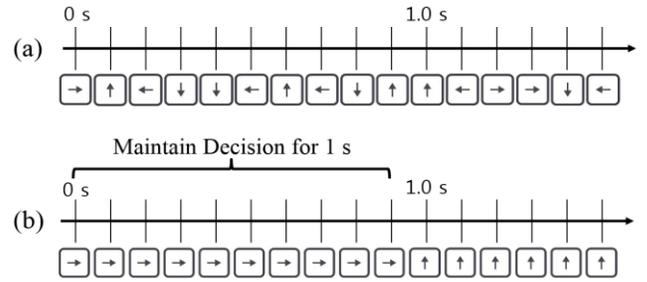

Figure 4. Examples of (a) regular move decisions and (b) maintaining decisions for 1 second

Figure 4 shows how the method works with an example. If the agent selects a move action, it skips the move decision for the following *n-1* time steps. This means that the agent maintains the same move decision for *n* steps in total. Note that our method has different purpose from frame skip technique (Mnih *et al*. 2015) in Atari domain. Frame skip technique was introduced for simulator's efficiency. However, we cannot just skip the frames because skill decisions must still be made. Although we could not enjoy advantage in the simulator's efficiency, maintaining move still facilitates training and this is solely because maintaining move decision increases the influence of a single move decision, as we will confirm with experiments. In this sense, maintaining move rather can be viewed as 'amplifying advantage' from (Mladenov *et al.* 2019).

## Experiments

### Implementation Details

#### Network
The network is composed of LSTM-based architecture which has four heads with a shared state representation layer. Each head consists of $\pi_{skill}$, $Q_{skill}$, $\pi_{move,target}$ and $Q_{move,target}$. $Q_{skill}$ and $Q_{move,target}$ are used for the gradient update of $\pi_{skill}$ and $\pi_{move,target}$, respectively. Before the network output goes into the softmax layer, a Boolean vector indicating the availability of each skill operates to make the output of unavailable skill to negative infinity.

#### Algorithm
We used actor-critic off-policy learning algorithm (Wang *et al.* 2017). It enables us to deal with policy lag between the simulators and learner through truncated importance sampling. Moreover, we could also use the advantages of stochastic policy, which responds more stably to changes in the environment due to smooth policy updates and works well in the domain of games like rock-scissors-paper where deterministic policy is vulnerable to exploitation. For this specific algorithm, both $\pi_{skill}$ and $\pi_{move,target}$ are updated in an alternating manner with following gradient:

$$g_t^{acer} = \bar{\rho}_t \nabla_\theta log \pi_\theta(a_t|x_t)[Q^{ret}(x_t, a_t) - V_{\theta_v}(x_t)] +$$
$$\mathbb{E}_{a \sim \pi}\left(\left[\frac{\rho_t(a) - c}{\rho_t(a)}\right]_+ \nabla_\theta log \pi_\theta(a|x_t)[Q_{\theta_v}(x_t, a) - V_{\theta_v}(x_t)]\right),$$

where $\bar{\rho}_t = \min\{c, \rho_t\}$ with behavior policy $\mu$ and importance sampling ratio $\rho_t = \frac{\pi(a_t|x_t)}{\mu(a_t|x_t)}$. $[x]_+ = x \text{ if } x > 0$ and zero otherwise.

**Learning System**

In total, there are three learning processes with each learning process consisting of a learner and 100 simulators. Each learning process is largely similar to that proposed by Horgan *et al.* (2018). The final agent is trained for two weeks, which is equivalent to four years of game play.

## Effect of Self-Play Curriculum with Three Styles

To demonstrate the effects of the proposed self-play curriculum, we trained agents with and without the proposed curriculum. A baseline agent was trained with the vanilla self-play curriculum without any style-related rewards (only win reward and HP reward were included) and a pool of past selves was used. Meanwhile, three agents with different styles were trained with the self-play curriculum using the shared pool that we proposed. Our aggressive, balanced and defensive agents[2] then played 1,000 matches each against the baseline agent to measure the performance. As shown in Table 3, the agents that followed the learnings from our curriculum outperformed the baseline agent.

Next, we conducted an ablation study to observe how the shared pool helps generalization. We wanted to confirm whether an agent would be able to deal with opponents of unseen style, when it experienced only a limited range of opponents during training. Thus, we created three styles of agents trained in exactly the same manner, except that they had their own independent opponent pools. We denote the three types of agents using shared pools as $\pi_{sh}^{agg}$, $\pi_{sh}^{bal}$, and $\pi_{sh}^{def}$, and three type of agents using independent pools as $\pi_{ind}^{agg}$, $\pi_{ind}^{bal}$, and $\pi_{ind}^{def}$. All of six agents were trained for 5M steps (equivalent to 6 days) each.

Our assumption is that the agent trained with the shared pool is more robust when it faces opponents it has never encountered. Thus, we compared the win rate of $\pi_{sh}^{agg}$ vs. $\{\pi_{ind}^{bal}, \pi_{ind}^{def}\}$ and $\pi_{ind}^{agg}$ vs. $\{\pi_{ind}^{bal}, \pi_{ind}^{def}\}$. This experimental setting is based on three key ideas. First, $\pi_{sh}^{agg}$ and $\pi_{ind}^{agg}$ have the same training settings except for sharing the pool. Second, $\pi_{sh}^{agg}$ and $\pi_{ind}^{agg}$ are evaluated against the same opponents. Finally, although $\pi_{sh}^{agg}$ has encountered other styles from its pool, it has not confronted $\{\pi_{ind}^{bal}, \pi_{ind}^{def}\}$, for they were trained using independent opponent pools. If our assumption is correct, $\pi_{sh}^{agg}$ should have a higher winning rate. It is to be noted that $\pi_{ind}^{bal}$ and $\pi_{ind}^{def}$ are not a single model, but 10 models each sampled at the same fixed intervals from their pools. We then conducted the same experiments for the remaining two styles; the results are presented in Table 4. As shown in the table, agents trained with shared pool outperform their counterparts.

---

[2] We measured how the average game length differs for each style because game length is a good proxy for assessing the degree of defensiveness of

|  | Aggressive | Balanced | Defensive | Average |
|---|---|---|---|---|
| Vs. Baseline | 59.5% | 63.8% | 63.2% | 62.2% |

Table 3: Win rate of three style of agents against baseline (1,000 games each)

|  | Aggressive | Balanced | Defensive | Average |
|---|---|---|---|---|
| Shared | 64.8% | 79.6% | 75.3% | 73.6% |
| Ind. | 64.7% | 72.1% | 56.5% | 64.4% |

Table 4: Generalization performance of three styles of agents for both with and without shared pool (7,000 games each)

Based on the data in Table 4, the effect of using a shared pool is marginal in the case of aggressive agents. It indicates that the strategy spaces in which trainings take place are similar whether or not various opponents are provided. This is related to the nature of fighting games in which one side should fight back if the other side approaches and initiates a brawl. Thus, in the case of an aggressive agent that attacks consistently, there is a little difference in the experience regardless of the diversity of the opponent's fighting style.

## Effect of Discarding Passive "No-op"

As discussed in the previous section, the "no-op" decision may be either active or passive. We conducted an experiment to investigate the effect of discarding such passive "no-op" data from learning. The sparring partner for the experiment was the built-in BAB AI, with a performance comparable to the top 20% of the players. We measured how fast agents learned to defeat it, and the results are shown in Figure 5 (a). If "no-op" ticks are discarded from the learning data, the win rate reaches 80% after 70k steps, whereas 170k steps are required when "no-op" ticks are included. The amount of time steps required to reach 90% win rate was reduced to half when passive "no-op" data was skipped.

an agent's game play. The results were as follows: 66.6 sec for the aggressive, 91.7 sec for the balanced, and 179.9 sec for the defensive agent.

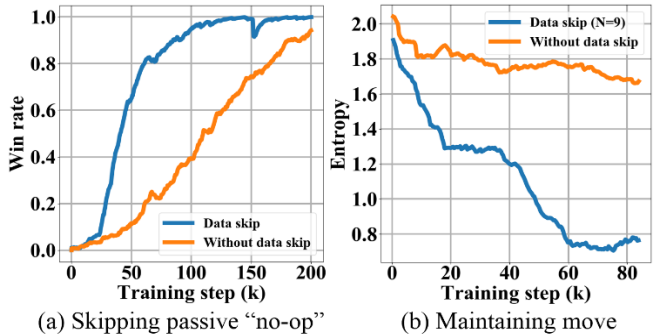

Figure 5. Results of data skipping experiments

(a) Skipping passive "no-op"  (b) Maintaining move

This experiment confirms that the training performance is improved by discarding passive "no-op" from the learning data.

**Effect of the Maintaining Move**

To examine the effect of the maintaining move, we developed two learning processes, with both processes involving learning on a self-play basis. One process makes a moving decision at every time step, while the other makes a moving decision and sends the same decision for 9 more times in a row. We measured the entropy of the move policy to observe the effects. Entropy of the move policy for a given state $s_t$ is as follows.

$$H(s_t) = -\sum \pi_{move}(s_t) * \log \pi_{move}(s_t)$$

Generally, entropy gradually decreases as learning progresses. Figure 5 (b) shows that the entropy declines faster if the technique is applied. A noticeable difference was also observed in the quality of movement which the agent learned. Before the technique was applied, the agent did not make any improvement from random motion, but it learned to approach and retreat with data skip.

The longer the decision was repeated, the agent's reaction became less immediate, but the agent moved more consistently. In this case, we tested 1, 3, 5, 10 ticks for maintaining time. 10 ticks (equivalent to 1 s) yielded the best performance.

## Pro-Gamer Evaluation

This section will address the results of both the pre-test and the Blind Match, and conditions to ensure fairness for human players.

### Conditions for Fairness
**Reaction Time**
When humans confront an AI in a real-time fighting game, the most important factor that affects the result is the reaction time. Humans require some time to recognize the skill used by the opponent and to press a button by moving his/her hand. We applied an average of 230 ms of delay for the decision of an AI to be reflected in the game, so that the AI does not have an advantage. This amount of delay corresponds to the average reaction time of professional players in BAB.

**Classes and Skill Set**
There are 11 classes in B&S, and each class has unique characteristics. Since there exists relative superiority among classes, we fixed the class of both AI and pro-gamer as "Destroyer". Destroyer is a class that has an infighting style and steadily appears in the B&S world championship. Additionally, AI's and pro-player's skill trees were set as identical to ensure a fair match. The skill tree was chosen to match what the majority of users selected, based on the BAB user statistics.

### Evaluation Results

We invited two prominent pro-gamers, Yuntae Son (GC Busan, Winner of 2017 B&S World Championship), and Shingyeom Kim (GC Busan, Winner of 2015 and 2016 B&S World Championship), to test our agents before the Blind Match. Note that the total number of games played is different for each style because the testers can play as many games as they want for each style. After the pre-test, we went for the Blind Match of 2018 World Championship. Our agents had matches against three pro-gamers: Nicholas Parkinson (EU), Shen Haoran (CHN), and Sungjin Choi (KOR). The video recording of the game highlights can be found at https://goo.gl/7VUTzV.

The results of both the pre-test and the Blind Match are provided in Table 5. As can be seen from the table, the aggressive agent dominated the game, while the other two types of agents had rather intense games. According to the interview after the pre-test, we found that this was partly because human players often need some breaks between fights, but the aggressive agent does not permit humans to have breaks between battles; rather, the attacks are continuous.

|  | Aggressive | Balanced | Defensive |
|---|---|---|---|
| Pro-Gamer 1 | 5-1 | 2-1 | 1-2 |
| Pro-Gamer 2 | 4-0 | 2-4 | 4-1 |
| Blind Match | 2-0 | 1-2 | 0-2 |
| Total | 11-1 (92%) | 5-7 (42%) | 5-5 (50%) |

Table 5: Final score of AI vs. Human

## Conclusion

Using deep reinforcement learning, we created AI agents that competed evenly with professional players in a 3D real-time fighting game. To accomplish this, we proposed a method to guide the fighting style with reward shaping. With three styles of agents, we introduced a novel self-play curriculum to enhance generalization performance. We also proposed data-skipping techniques to improve data efficiency and enable efficient exploration. Consequently, our agents were able to compete with the best BAB pro-gamers in the world. The proposed training methods are generally applicable to other fighting games.